\documentclass[a4paper,oneside,bibliography=totoc]{scrbook}

\usepackage[utf8]{inputenc}
\usepackage{graphicx}
\DeclareGraphicsExtensions{.pdf,.png,.jpg}
\usepackage{latexsym}
\usepackage{amsmath}
\usepackage{amssymb}
\usepackage{tabularx}
\usepackage{booktabs}
\usepackage{algorithm} 
\counterwithin{algorithm}{chapter} 
\usepackage{algorithmic}
\usepackage{csquotes}

\usepackage[usenames,dvipsnames]{xcolor}
\usepackage[colorlinks,citecolor=Green]{hyperref} 
\usepackage{lipsum} 

\usepackage{tikz} 
\tikzset{
  process/.style = {rectangle, draw, minimum height=2em, text width=12em, align=center},
  arrow/.style = {thick, ->, >=stealth}
} 

\usepackage{multicol}

\usepackage{subcaption}

\usepackage{natbib}
\setcitestyle{authoryear,round,semicolon,aysep={},yysep={,}} \let\cite\citep

\usepackage[toc,page]{appendix}
\begin{document}


\title{\textbf{Enhancing mortality prediction in cardiac arrest ICU patients through meta-modeling of structured clinical data from MIMIC-IV}}

\author{
  Philipp Kellmeyer\textsuperscript{1,*} \and
  Nursultan Mamatov\textsuperscript{1}
}

\date{
  \small
  \textsuperscript{1}Data and Web Science Group, School of Business Informatics and Mathematics, University of Mannheim, Germany\\[4pt]
  *Corresponding author: \texttt{philipp.kellmeyer@uni-mannheim.de}\\[8pt]
}

\maketitle

\chapter{Abstract}

Accurate early prediction of in-hospital mortality in intensive care units (ICUs) is essential for timely clinical intervention and efficient resource allocation. This study develops and evaluates machine learning models that integrate both structured clinical data and unstructured textual information, specifically discharge summaries and radiology reports, from the MIMIC-IV database. We used LASSO and XGBoost for feature selection, followed by a multivariate logistic regression trained on the top features identified by both models. Incorporating textual features using TF-IDF and BERT embeddings significantly improved predictive performance. The final logistic regression model, which combined structured and textual input, achieved an AUC of 0.918, compared to 0.753 when using structured data alone, a relative improvement 22\%. The analysis of the decision curve demonstrated a superior standardized net benefit in a wide range of threshold probabilities (0.2-0.8), confirming the clinical utility of the model. These results underscore the added prognostic value of unstructured clinical notes and support their integration into interpretable feature-driven risk prediction models for ICU patients.

\begingroup%
\hypersetup{hidelinks}
\tableofcontents%
\endgroup


\mainmatter

\chapter{Introduction}

\section{Background and Motivation}

Cardiac arrest (CA) remains one of the most critical and fatal emergencies worldwide. In the United States alone, over 350{,}000 out-of-hospital and about 292{,}000 in-hospital cardiac arrests occur annually, with survival to discharge rarely exceeding 25\%~\citep{benjamin2018heart,holmberg2019annual,andersen2019adult}. Even among survivors, post-cardiac arrest syndrome often leads to neurological and multi-organ complications~\citep{nolan2007outcome}.  

Patients resuscitated after CA are typically admitted to intensive care units (ICUs) for advanced monitoring and life support. Yet mortality in this group remains high~\citep{sun2023prediction}. Early identification of high-risk patients is vital both for optimizing treatment decisions and for prioritizing limited ICU resources.

\section{Problem Statement}

While machine learning (ML) models based on structured data—vital signs, labs, demographics, and comorbidities—have improved mortality prediction~\citep{sun2023prediction}, they often ignore unstructured text such as clinical notes and radiology reports. These narratives contain rich contextual cues about patient condition and clinical reasoning. Moreover, many ML models lack interpretability~\citep{ennab2022designing}, limiting their acceptance in clinical workflows. To enable responsible AI, both predictive accuracy and transparency must be improved through multimodal data integration.

\section{Objectives and Research Questions}

This study develops interpretable models that integrate structured and textual information from the MIMIC-IV database to predict in-hospital mortality in ICU-admitted CA patients. The main research questions are:

\begin{itemize}
    \item \textbf{RQ1:} Can the structured-data-based mortality model of \citet{sun2023prediction} be replicated and validated?
    \item \textbf{RQ2:} Does adding unstructured clinical notes (discharge summaries and radiology reports) improve predictive performance and clinical utility?
\end{itemize}

\section{Significance and Contribution}

This thesis bridges quantitative modeling and clinical narratives through a multimodal framework that fuses structured EHR variables with text embeddings derived from TF-IDF + SVD and BioBERT. Using feature selection via LASSO and XGBoost and interpretable logistic regression, it demonstrates that unstructured data enhance prediction and interpretability.

Structured-data-only models achieved an AUC of 0.75, while adding textual features improved performance to 0.92 and expanded the range of clinical benefit in decision curve analysis. The framework remains fully interpretable through explicit regression coefficients and binary indicators of note presence.  

Overall, this work shows that combining structured and unstructured data can yield transparent and high-performing predictive tools for post–cardiac arrest ICU care.

---

\chapter{Literature Review}
\label{ch:related_work}

\section{Overview}
\label{sec:overview}

Machine learning is increasingly applied to intensive care medicine, with many studies focusing on mortality prediction from structured EHR data. However, textual records—such as discharge summaries and radiology reports—contain critical clinical insights that remain underused. This chapter reviews prior work on structured-data models, text-based prediction, multimodal fusion, and interpretability in healthcare AI.

\section{ICU Mortality Prediction Using Structured Data}
\label{sec:structured_data}

Structured variables such as lab results, vitals, and demographics form the basis of most ICU mortality models. \citet{sun2023prediction} trained ML algorithms on MIMIC-IV data and achieved an AUC of 0.79 using XGBoost. Similarly, \citet{harutyunyan2019multitask} benchmarked four prediction tasks on MIMIC-III, showing the benefit of sequence models for capturing temporal ICU dynamics. These works demonstrate the predictive potential of structured EHR data but omit unstructured contextual information.

\section{Text-Based Modeling and ClinicalBERT}
\label{sec:textual_data}

Unstructured clinical text captures subtleties such as clinician judgment, disease evolution, and treatment rationale. \citet{huang2019clinicalbert} introduced ClinicalBERT, a transformer model pretrained on MIMIC-III notes, which outperformed conventional methods for 30-day readmission prediction and provided interpretable attention mechanisms. Such domain-adapted language models enable robust and meaningful text representations for downstream clinical tasks.

\section{Multimodal Learning and Data Integration}
\label{sec:multimodal_models}

Combining structured and unstructured data can significantly enhance prediction. \citet{rajkomar2018scalable} demonstrated end-to-end deep learning on raw EHR data—including notes—achieving AUCs above 0.93 for in-hospital mortality. More recent hybrid approaches fuse structured features with embeddings from models like ClinicalBERT or TF-IDF + SVD. However, interpretability often remains limited. This thesis addresses that gap by using transparent meta-modeling (LASSO, XGBoost, logistic regression) to balance accuracy and explainability.

\section{Clinical Context: Post–Cardiac Arrest Outcomes}
\label{sec:clinical_context}

Post-cardiac arrest patients represent a uniquely high-risk population. \citet{dutta2022brain} identified brain natriuretic peptide (BNP) as a discriminator of poor neurological outcomes, while \citet{andersson2021predicting} showed that models combining biomarkers and clinical variables achieved AUROCs above 0.94. These findings highlight the value of multimodal information for outcome prediction and motivate similar integration in mortality modeling.

\section{Interpretability and Responsible AI}
\label{sec:interpretability}

For adoption in clinical practice, predictive models must be explainable. \citet{lundberg2017unified} introduced SHAP values for model-agnostic feature attribution, offering intuitive interpretability. \citet{si2021deep} further emphasized multimodal patient representations as essential for capturing complex health states. Building on these principles, this thesis employs interpretable regression and explicit feature selection rather than opaque black-box models.

\section{Summary}
\label{sec:summary}

Existing studies show that structured EHR data provide solid predictive baselines, unstructured text adds complementary context, and multimodal integration yields superior results. Yet few frameworks combine these sources within an interpretable structure. This thesis advances the field by developing a transparent multimodal model for ICU mortality prediction after cardiac arrest, integrating structured metrics with clinically meaningful textual representations.

\chapter{Data Description}
\label{ch:data_description}

\section{Overview}
\label{sec:overview}

This study uses two publicly available datasets from the Medical Information Mart for Intensive Care (MIMIC) project: \textit{MIMIC-IV} and \textit{MIMIC-IV-Note}. Both are developed and maintained by the Laboratory for Computational Physiology at MIT and accessed through the PhysioNet repository after completing data use training and certification.

MIMIC-IV provides structured electronic health record (EHR) data, while MIMIC-IV-Note offers de-identified clinical text. Together, they form a comprehensive multimodal resource that integrates quantitative patient measurements with narrative documentation.

\section{Structured Data (MIMIC-IV)}
\label{sec:structured_data_description}

MIMIC-IV is a large-scale, de-identified database containing health records from patients admitted to the Beth Israel Deaconess Medical Center between 2008 and 2022. It includes over 364,000 unique individuals and 94,000 ICU stays. The dataset is organized into two modules:  
\begin{itemize}
    \item \texttt{hosp}: hospital-wide data, including demographics, diagnoses (ICD codes), labs, prescriptions, and microbiology results;
    \item \texttt{icu}: ICU-specific time-stamped events such as vital signs, medication infusions, and interventions.
\end{itemize}

Each record is linked through hierarchical identifiers allowing patient-level integration across tables. All timestamps are anonymized by patient-specific temporal shifts to future years (2100–2200) while maintaining internal consistency. Anchor variables (\texttt{anchor\_age}, \texttt{anchor\_year}) enable approximate temporal alignment without revealing true dates.

\begin{figure}[htbp]
    \centering
    \includegraphics[width=0.9\linewidth]{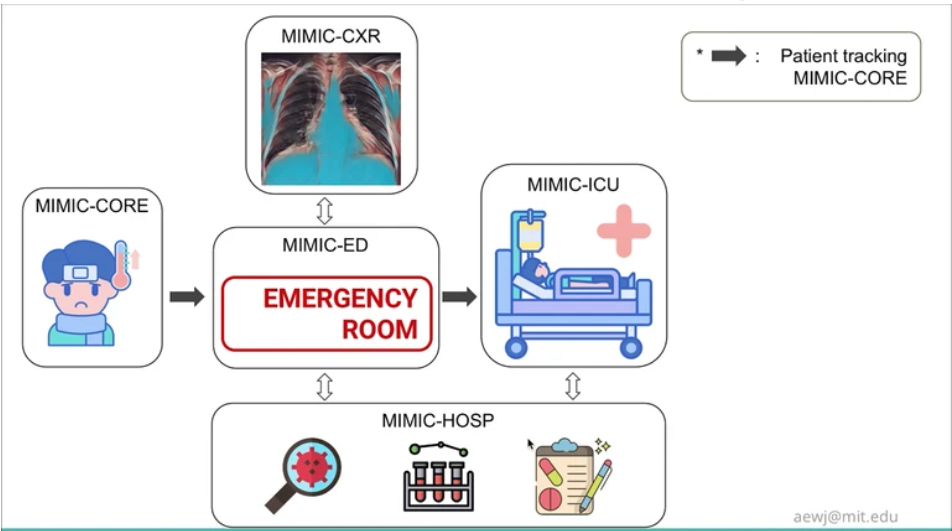}
    \caption{Simplified flow of data capture across MIMIC-IV modules, adapted from~\cite{harutyunyan2019multitask}.}
    \label{fig:mimic_core_flow}
\end{figure}

Patient data flow is systematically captured: from emergency department admission to ICU stay and eventual discharge. The modular structure ensures that hospital- and ICU-level data remain synchronized for each patient encounter (Fig.~\ref{fig:mimic_core_flow}).

\section{Unstructured Data (MIMIC-IV-Note)}
\label{sec:text_data_description}

The \textit{MIMIC-IV-Note} extension includes more than 2.6 million de-identified clinical notes—approximately 2.3 million radiology reports and 330,000 discharge summaries — covering over 230,000 patients. Each note is linked to the structured dataset via \texttt{subject\_id} and \texttt{hadm\_id}.

\textbf{Discharge summaries} describe admission reasons, hospital course, and discharge outcomes, while \textbf{radiology reports} follow standardized sections such as indication, findings, and impression. These notes capture clinical reasoning and contextual detail often missing from structured fields.

De-identification follows HIPAA Safe Harbor guidelines using a hybrid rule-based and neural-network pipeline, with identified protected health information replaced by underscores (\texttt{\_\_\_}). Manual audits report over 99\% sensitivity for PHI removal~\cite{johnson2020deidentification}.

\section{Data Integration}
\label{sec:data_integration}

Structured and unstructured data were merged using shared identifiers (\texttt{subject\_id}, \texttt{hadm\_id}, \texttt{stay\_id}). This linkage aligns physiological measurements, laboratory values, comorbidities, and interventions with corresponding textual narratives such as discharge or radiology notes.  

Consistent temporal offsets ensure chronological coherence within each patient stay. The resulting dataset enables multimodal learning that combines objective physiological data with subjective clinical observations—providing a rich foundation for mortality prediction, risk stratification, and other critical care applications.

\chapter{Non-Textual Data Preprocessing}
\label{ch:data_preprocessing}

\section{Cohort Definition and Filtering}
\label{sec:cohort_filtering}

This study used MIMIC-IV v3.0 to construct a cohort of adult patients who experienced cardiac arrest (CA) and were admitted to the intensive care unit (ICU). The objective was to create a clean, reproducible dataset suitable for mortality prediction.

\subsection*{Cohort Selection}
Three primary tables were used: \texttt{diagnoses\_icd}, \texttt{patients}, and \texttt{icustays}, containing diagnostic codes, demographics, and ICU stay metadata.  
CA patients were identified by ICD-9 and ICD-10 codes (\texttt{4275}, \texttt{I46}, \texttt{I462}, \texttt{I468}, \texttt{I469}), yielding 2,653 admissions (2,608 unique patients). Duplicates by \texttt{hadm\_id} were removed to retain one record per admission.

To ensure ICU relevance, the filtered cohort was merged with the \texttt{icustays} table via \texttt{subject\_id} and \texttt{hadm\_id}, resulting in 2,881 ICU stays from 2,307 patients. Only the first ICU stay per patient was kept, representing the index event and avoiding bias from readmissions.

\subsection*{Eligibility and Outcome Definition}
Only adult patients (age $\geq$18) were included, determined via the \texttt{anchor\_age} field.  
In-hospital mortality was defined using the \texttt{deathtime} and \texttt{dischtime} fields from the \texttt{admissions} table:

\[
\texttt{in\_hospital\_death} =
\begin{cases}
1, & \text{if } \texttt{deathtime} \le \texttt{dischtime} \\
0, & \text{otherwise}
\end{cases}
\]

This yielded 1,207 deaths (52.3\%) and 1,100 survivors (47.7\%). The resulting cohort of 2,307 first-time ICU admissions formed the analytical base for subsequent modeling.

\section{Vital Signs: Extraction and Harmonization}
\label{sec:vital_signs}

Vital sign events were extracted from the \texttt{chartevents} table, restricted to the first 24 hours of ICU admission. The selected variables were heart rate (HR), systolic/diastolic/mean blood pressure (SBP, DBP, MBP), respiratory rate (RR), body temperature (BT), and oxygen saturation (SpO$_2$).

\subsection*{Cleaning and Aggregation}
Outliers were removed using physiological limits and unit inconsistencies (e.g., Celsius vs Fahrenheit) were corrected.  
Each variable’s mean, minimum, and maximum within 24 hours were computed per patient, then merged with demographic and clinical metadata.

\subsection*{Measurement Harmonization}
Temperature values were unified using:
\[
\text{BT}_F = \text{BT}_C \times \frac{9}{5} + 32
\]
Arterial and non-invasive blood pressures were aligned via regression-based conversions~\citep{liu2014comparison} and averaged:
\[
\text{MBP} = \frac{\text{SBP} + 2 \times \text{DBP}}{3}
\]
This standardization maximized data coverage while maintaining physiological realism.

\subsection*{Missingness and Imputation}
Most patients had complete vital data (2,301/2,307). Missing values were imputed as follows:
\begin{itemize}
    \item BT (13.3\% missing): reserved for multiple imputation.  
    \item HR, DBP: mean imputation (low missingness).  
    \item SBP, MBP, RR, SpO$_2$: median imputation (mild skew).  
\end{itemize}

Summary statistics and visual distributions are available in Appendices~\ref{app:missing_data} and~\ref{app:vital_figures}.  
The processed vital signs dataset is fully harmonized and ready for analysis.

\section{Laboratory Data Processing}
\label{sec:labevents}

Laboratory measurements were extracted from the \texttt{labevents} table for all cohort members, filtered by hospital admission (\texttt{hadm\_id}) and limited to the first 24 ICU hours. The resulting subset comprised 320,962 lab observations across 2,239 patients.

\subsection*{Variable Selection and Cleaning}
Seventeen clinically relevant lab tests were retained: HCT, HB, PLT, WBC, PT, INR, creatinine, BUN, glucose, potassium, sodium, calcium, chloride, anion gap, bicarbonate, lactate, and pH. Implausible values (e.g., WBC $<$1 or $>$50, glucose $>$600 mg/dL, lactate $>2$0 mmol/L) were removed. Each variable’s mean, min, and max were aggregated to patient level and merged with metadata.

\subsection*{Handling Missingness}
Of the 2,307 patients, 71 had no lab data. Missingness rates ranged from 4\%–6\% for most variables, with lactate (19\%) and pH (17.6\%) missing more frequently.  
Imputation followed three rules:
\begin{itemize}
    \item Median imputation for skewed variables (e.g., creatinine, glucose).  
    \item Mean imputation for $<$5\% missing (e.g., sodium, bicarbonate).  
    \item No imputation for $>$10\% missing (lactate, pH) — handled later via multiple imputation.
\end{itemize}

Distributions and skewness summaries appear in Appendix~\ref{app:lab_figures}.  
The final lab dataset includes 17 cleaned and aligned variables for 2,307 ICU stays.

\section{Comorbidities and Treatment Indicators}
\label{sec:comorbidities_treatment}

\subsection*{Comorbidity Extraction}
Major pre-existing conditions were identified from the \texttt{diagnoses\_icd} table using standard ICD mappings.  
Binary indicators were created for five common comorbidities: hypertension, heart failure, myocardial infarction, diabetes, and COPD.  
These were merged with the master dataset using \texttt{hadm\_id}.

\subsection*{Treatment Indicators}
To reflect early ICU intervention intensity, three treatment variables were derived from time-stamped \texttt{procedureevents} and \texttt{inputevents} tables:
\begin{itemize}
    \item Mechanical ventilation (within 24h): received by 59.7\% of patients.  
    \item Epinephrine: 2.6\% of patients.  
    \item Dopamine: 2.1\% of patients.  
\end{itemize}
Each was encoded as a binary flag and merged using \texttt{stay\_id}.

\section{Neurological Assessment: Glasgow Coma Scale}
\label{sec:marking_systems}

Neurological status was represented by the Glasgow Coma Scale (GCS) components—Eye (1–4), Verbal (1–5), and Motor (1–6)—extracted from the \texttt{chartevents} table.  
Mean values during the ICU stay were computed, and missingness was minimal ($<2.5\%$ per component). Mean or median imputation was used based on skewness.  
A total score (\texttt{GCS\_Total} = Eye + Verbal + Motor, range 3–15) was then derived, providing an interpretable measure of consciousness severity.  
Distribution plots are provided in Appendix~\ref{app:gcs_figures}.

\section{Multiple Imputation and Final Dataset}
\label{sec:missing_data}

After merging all structured components—vitals, labs, comorbidities, GCS, and treatments—eight variables exhibited moderate missingness (5–19\%).  
We applied multiple imputation via chained equations (MICE) using the \texttt{IterativeImputer} from \texttt{scikit-learn}~\citep{pedregosa2011scikit} with $m=5$ iterations.  
Parameter estimates were combined using Rubin’s Rules~\citep{lee2014introduction} to account for within- and between-imputation variance:

\[
\hat{\beta}_{\text{MI}} = \frac{1}{m} \sum_{i=1}^{m} \hat{\beta}_i, \quad
T = V + \left(1+\frac{1}{m}\right)B
\]

The resulting complete dataset retained all 2,307 patients without casewise deletion, ensuring consistency for subsequent modeling and analysis.

\chapter{Textual Data Preprocessing}
\label{ch:textual_preprocessing}

Unstructured clinical notes provide rich insights into patient status, capturing reasoning, context, and temporal evolution often missing from structured data. This study incorporated two types of notes from MIMIC-IV—discharge summaries and radiology reports—transforming them into numerical features using both classical and transformer-based NLP methods.

\section{Text Source Selection and Filtering}
\label{sec:text_source_filtering}

All notes were linked to hospital admissions via \texttt{hadm\_id} and filtered to include only those associated with the 2,307 ICU patients in our structured cohort.  

\textbf{Discharge summaries} were retained when available, yielding 1,618 documents (70.1\% coverage).  
\textbf{Radiology reports} often had multiple entries per admission; therefore, only the earliest report per stay was selected to ensure temporal relevance. This resulted in 1,639 reports (71.0\% coverage).

This filtering preserved one text sample per modality per patient, minimizing redundancy and ensuring consistency with the structured dataset.

\section{Text Normalization and Feature Extraction}
\label{sec:text_vectorization}

Two complementary pipelines were developed:

\begin{itemize}
    \item \textbf{TF-IDF Representation:} Texts were lowercased and stripped of punctuation, digits, and English stopwords. A 500-term vocabulary was constructed separately for discharge and radiology notes to reflect their differing content. Sparse document-term matrices were then generated.
    \item \textbf{BioBERT Embeddings:} Using the pretrained BioBERT model~\citep{alsentzer2019publicly}, each note was encoded into a 768-dimensional dense vector via mean-pooling across the final hidden layer. These embeddings capture biomedical context and semantic nuance.
\end{itemize}

\section{Dimensionality Reduction}
\label{sec:text_reduction}

To reduce dimensionality and redundancy:

\begin{itemize}
    \item \textbf{TF-IDF features} were compressed using TruncatedSVD, retaining 80\% of variance—136 components for discharge notes and 198 for radiology notes (Appendix~\ref{app:text_figures_tfidf}).
    \item \textbf{BioBERT embeddings} were reduced using PCA, preserving 90\% of variance—113 and 115 components, respectively (Appendix~\ref{app:text_figures_bert}).
\end{itemize}

These transformations improved efficiency while maintaining interpretive fidelity across text types.

\section{Integration with Structured Data}
\label{sec:text_merge}

Reduced textual features were merged with structured variables via \texttt{hadm\_id}. The merged dataset added 562 text-derived columns.

Coverage varied: 689 patients lacked discharge summaries, 668 lacked radiology notes, and 712 had at least one missing modality.  
Missing text embeddings were replaced with \textbf{zero vectors}, reflecting the true absence of semantic content rather than uncertainty.  
Zero imputation is consistent with the statistical structure of both feature types—TruncatedSVD TF-IDF and PCA BioBERT embeddings are approximately zero-centered.  

To enhance interpretability, binary indicators were added, allowing models to distinguish between “no document” and “zero-valued” text features.

\section{Summary}
\label{sec:text_summary}

This chapter detailed the transformation of MIMIC-IV discharge and radiology notes into structured embeddings for multimodal modeling. Text data were filtered for cohort alignment, vectorized via TF-IDF and BioBERT, dimensionally reduced for efficiency, and merged with the structured dataset. Zero imputation and binary note indicators preserved interpretability while maintaining data integrity.  

The resulting multimodal dataset integrates structured clinical variables with context-rich text representations, forming the basis for the predictive modeling described in the next chapter.

\chapter{Modeling with Structured Data}
\label{chap:structured_modeling}

This chapter presents the development and evaluation of predictive models for in-hospital mortality using structured features from the MIMIC-IV dataset. The workflow combined feature selection, logistic regression modeling, and comprehensive model assessment through discrimination, calibration, and decision analysis.

\section{Predictor Selection via Regularized and Ensemble Methods}
\label{sec:feature_selection_methods}

To identify robust predictors while maintaining interpretability, two complementary feature selection techniques were applied: LASSO (Least Absolute Shrinkage and Selection Operator) and XGBoost (Extreme Gradient Boosting). LASSO offers a linear and interpretable approach that enforces sparsity, while XGBoost captures nonlinear feature interactions and assigns importance scores through gradient boosting. Together, they provide a balanced view of linear and nonlinear relationships within the structured data.

\subsection*{LASSO Logistic Regression}

LASSO extends logistic regression by introducing an $\ell_1$ penalty that shrinks less relevant coefficients toward zero, effectively performing feature selection. The optimal penalty strength (\(\lambda\)) was determined via 10-fold cross-validation using scikit-learn’s SAGA solver to minimize the binomial deviance. Features with non-zero coefficients in the final model were retained as candidate predictors.

\subsection*{XGBoost Feature Importance}

XGBoost is an ensemble of decision trees optimized through gradient boosting. It sequentially adds trees that correct previous residual errors while regularization prevents overfitting. Model parameters (\texttt{max\_depth}=3, \texttt{learning\_rate}=0.05, \texttt{n\_estimators}=100, \texttt{subsample}=0.8) were tuned for stability and generalization. Feature relevance was ranked using the “gain” metric, reflecting each variable’s contribution to model improvement.

\subsection*{Feature Set Consolidation}

Outputs from both methods were combined to form a unified candidate set:
\begin{enumerate}
    \item Predictors with non-zero coefficients from the LASSO model.
    \item The most important features from XGBoost (based on gain ranking).
\end{enumerate}

Merging these yielded 20 unique predictors, which were further screened through univariate logistic regression to confirm their association with in-hospital mortality (\(p < 0.05\)). This consolidated set formed the foundation for all subsequent multivariable modeling.

\section{Initial Screening Using Univariate Logistic Regression}
\label{sec:univariate_lr}

Each variable from the consolidated feature set was individually assessed using univariate logistic regression. Out of the 20 candidates, 18 were found to have a statistically significant association with in-hospital mortality (\(p < 0.05\)). The full results, including coefficients and p-values, are presented in Table~\ref{table:uni_logit_results}.

\begin{table}[ht]
\centering
\caption{Univariate Logistic Regression Results for Combined Features}
\label{table:uni_logit_results}
\begin{tabular}{lrr}
\toprule
\textbf{Variable} & \textbf{Coefficient} & \textbf{P-value} \\
\midrule
Lactate                 & 0.2712  & $<0.0001$ \\
SBP                     & -0.0153 & $<0.0001$ \\
BUN                     & 0.0203  & $<0.0001$ \\
Chloride                & -0.0007 & 0.9253    \\
Bicarbonate             & -0.1177 & $<0.0001$ \\
HR                      & 0.0260  & $<0.0001$ \\
INR                     & 0.8393  & $<0.0001$ \\
received\_ventilation   & -0.0728 & 0.4717    \\
RR                      & 0.0990  & $<0.0001$ \\
copd                    & 0.2466  & 0.0440    \\
PT                      & 0.0809  & $<0.0001$ \\
Hemoglobin              & -0.1094 & $<0.0001$ \\
WBC                     & 0.0440  & $<0.0001$ \\
SpO$_2$                 & -0.1191 & $<0.0001$ \\
GCS\_Total              & -0.0941 & $<0.0001$ \\
pH                      & -5.9490 & $<0.0001$ \\
BT                      & -0.2304 & $<0.0001$ \\
anchor\_age             & 0.0107  & 0.0005    \\
AnionGap                & 0.1289  & $<0.0001$ \\
congestive\_heart\_failure & -0.2480 & 0.0145 \\
\bottomrule
\end{tabular}
\end{table}

\section{Multicollinearity Assessment}
\label{sec:vif_analysis}

Before fitting the multivariable logistic regression model, we assessed multicollinearity among the 18 significant predictors using the Variance Inflation Factor (VIF). High multicollinearity inflates coefficient variances and reduces interpretability~\cite{belsley2005regression}. 

VIF measures how much the variance of a predictor increases due to correlation with other variables and is computed as:
\[
\text{VIF}_j = \frac{1}{1 - R_j^2},
\]
where \( R_j^2 \) is obtained by regressing predictor \( X_j \) on all others. Values above 5 typically indicate problematic collinearity.

\textbf{Key Finding:} \texttt{INR} and \texttt{PT} showed extremely high VIFs ($>60$), confirming redundancy as both represent coagulation status. To improve model stability, \texttt{INR} was removed while \texttt{PT} was retained. All remaining predictors had acceptable VIFs (mostly near 1), indicating low collinearity and reliable coefficient estimates (Appendix~\ref{app:vif_table}).

\section{Multivariate Logistic Regression Models}
\label{sec:multivariate_models}

To identify independent predictors of in-hospital mortality, we built and compared three multivariate logistic regression models derived from different feature selection methods:

\begin{enumerate}
    \item \textbf{LASSO-based model:} 17 predictors from LASSO regularization; after univariate filtering (\(p<0.05\)), 16 were retained.
    \item \textbf{XGBoost-based model:} 17 top-ranked variables from XGBoost; after filtering and removing \texttt{INR} (collinear with \texttt{PT}), 15 predictors remained.
    \item \textbf{Full model:} Combined all significant predictors from both methods; \texttt{INR} removed, yielding 17 variables.
\end{enumerate}

All models were fit using \texttt{statsmodels.Logit}, providing regression coefficients, confidence intervals, and \(p\)-values for inference.

\vspace{0.5em}
\subsection*{LASSO-based Model}

The final LASSO model (16 features) achieved a pseudo-\(R^2\) of 0.198, indicating moderate explanatory power. Higher lactate, heart rate, INR, and age were associated with increased mortality, while higher GCS, temperature, hemoglobin, and SpO$_2$ were protective—consistent with known physiological markers.

\begin{table}[ht]
\centering
\caption{LASSO-Based Multivariate Logistic Regression Results}
\label{tab:lasso_logit}
\resizebox{\textwidth}{!}{
\begin{tabular}{lrrr}
\toprule
\textbf{Variable} & \textbf{Coefficient} & \textbf{P-value} & \textbf{Interpretation} \\
\midrule
Lactate & +0.113 & $<$0.001 & Higher lactate $\uparrow$ mortality risk \\
GCS\_Total & --0.066 & $<$0.001 & Lower GCS $\downarrow$ survival odds \\
anchor\_age & +0.017 & $<$0.001 & Older age $\uparrow$ mortality risk \\
HR & +0.021 & $<$0.001 & Higher heart rate $\uparrow$ mortality risk \\
BT & --0.158 & $<$0.001 & Lower temperature $\uparrow$ mortality risk \\
INR & +0.367 & $<$0.001 & Elevated INR $\uparrow$ mortality risk \\
SpO\textsubscript{2} & --0.056 & 0.007 & Lower oxygen saturation $\uparrow$ mortality risk \\
Hemoglobin & --0.082 & 0.002 & Lower hemoglobin $\uparrow$ mortality risk \\
BUN & +0.014 & $<$0.001 & Higher BUN $\uparrow$ mortality risk \\
WBC & +0.020 & 0.028 & Higher WBC count $\uparrow$ mortality risk \\
copd & +0.332 & 0.019 & COPD $\uparrow$ mortality risk \\
congestive\_heart\_failure & --0.383 & 0.002 & CHF $\downarrow$ survival odds \\
\bottomrule
\end{tabular}
}
\end{table}

\subsection*{XGBoost-based Model}

The XGBoost-derived model (15 predictors) achieved a pseudo-\(R^2 = 0.196\). Despite XGBoost’s nonlinear nature, its logistic regression counterpart remained interpretable, with nearly identical key predictors—lactate, GCS, HR, BUN, and BT—underscoring consistency across selection methods.

\begin{table}[ht]
\centering
\caption{XGBoost-Based Multivariate Logistic Regression Results}
\label{tab:xgb_logit}
\resizebox{\textwidth}{!}{
\begin{tabular}{lrrr}
\toprule
\textbf{Variable} & \textbf{Coefficient} & \textbf{P-value} & \textbf{Interpretation} \\
\midrule
Lactate & +0.112 & $<$0.001 & Higher lactate $\uparrow$ mortality risk \\
GCS\_Total & --0.066 & $<$0.001 & Lower GCS $\downarrow$ survival odds \\
anchor\_age & +0.018 & $<$0.001 & Older age $\uparrow$ mortality risk \\
HR & +0.021 & $<$0.001 & Higher heart rate $\uparrow$ mortality risk \\
BT & --0.160 & $<$0.001 & Lower temperature $\uparrow$ mortality risk \\
BUN & +0.015 & $<$0.001 & Higher BUN $\uparrow$ mortality risk \\
SpO\textsubscript{2} & --0.057 & 0.007 & Lower oxygen saturation $\uparrow$ mortality risk \\
Hemoglobin & --0.083 & 0.002 & Lower hemoglobin $\uparrow$ mortality risk \\
WBC & +0.022 & 0.019 & Higher WBC count $\uparrow$ mortality risk \\
congestive\_heart\_failure & --0.358 & 0.004 & CHF $\downarrow$ survival odds \\
\bottomrule
\end{tabular}
}
\end{table}

\subsection*{Full Model}

The combined model included 17 predictors and achieved pseudo-\(R^2 = 0.1981\), nearly identical to the LASSO model. It captured key physiological domains—hemodynamic (HR, SBP), respiratory (SpO$_2$), metabolic (lactate, BUN), hematologic (hemoglobin, PT), and comorbidities (COPD, CHF)—confirming their independent prognostic contributions.

\begin{table}[ht]
\centering
\caption{Final Multivariate Logistic Regression Results}
\label{tab:final_logit}
\resizebox{\textwidth}{!}{
\begin{tabular}{lrrr}
\toprule
\textbf{Variable} & \textbf{Coefficient} & \textbf{P-value} & \textbf{Interpretation} \\
\midrule
anchor\_age & +0.017 & $<$0.001 & Age $\uparrow$ mortality risk \\
Lactate & +0.114 & 0.001 & Lactate $\uparrow$ mortality risk \\
GCS\_Total & --0.066 & $<$0.001 & Lower GCS $\downarrow$ survival \\
HR & +0.021 & $<$0.001 & HR $\uparrow$ mortality risk \\
BT & --0.159 & $<$0.001 & Lower BT $\uparrow$ mortality risk \\
Hemoglobin & --0.082 & 0.003 & Lower hemoglobin $\uparrow$ mortality risk \\
SpO\textsubscript{2} & --0.057 & 0.006 & SpO$_2$ $\downarrow$ survival \\
PT & +0.035 & $<$0.001 & Longer PT $\uparrow$ mortality risk \\
BUN & +0.015 & $<$0.001 & BUN $\uparrow$ mortality risk \\
WBC & +0.020 & 0.027 & WBC $\uparrow$ mortality risk \\
copd & +0.329 & 0.020 & COPD $\uparrow$ mortality risk \\
congestive\_heart\_failure & --0.377 & 0.003 & CHF $\downarrow$ survival \\
\bottomrule
\end{tabular}
}
\end{table}

\subsection*{Model Comparison}

All three models yielded consistent results and effect directions, confirming robustness. Shared predictors—lactate, GCS, age, HR, and BUN—consistently emerged as dominant factors. The LASSO model achieved the highest pseudo-\(R^2\) (0.1983), closely followed by the full model (0.1981) and XGBoost model (0.1957). Overall, the LASSO model offered the best balance between parsimony, interpretability, and predictive power, making it suitable for clinical application.

\section{Model Evaluation}
\label{sec:evaluation}

The predictive performance of the structured data models was evaluated in terms of discrimination, calibration, and clinical utility. Discrimination was assessed using the Area Under the Receiver Operating Characteristic Curve (AUC-ROC) on the validation set (Figure~\ref{app:roc_structured}). The XGBoost-based model achieved the highest AUC (0.77), followed closely by the LASSO-based and full models, both at 0.75. For comparison, the conventional NEWS2 score yielded an AUC of 0.67, confirming its lower discriminative ability relative to machine learning–based models.

Calibration analysis examined how closely predicted probabilities matched observed outcomes (Figure~\ref{app:calibration_structured_models}). The LASSO and XGBoost models displayed good agreement with the ideal diagonal, indicating reliable risk estimation. Although the NEWS2 model exhibited lower discrimination, its calibration curve was well aligned, suggesting that it still provided reasonably accurate probability estimates.

Clinical utility was assessed through Decision Curve Analysis (DCA) across threshold probabilities from 0.01 to 0.99 (Figure~\ref{app:dca_structured_models}). All structured logistic models offered greater net benefit than “treat-all” or “treat-none” strategies, particularly within the clinically relevant threshold range of 0.2–0.8. The LASSO and XGBoost models demonstrated the highest benefit in intermediate-risk regions, where decision uncertainty is typically greatest.

To benchmark these models against a widely used clinical tool, we evaluated the National Early Warning Score 2 (NEWS2). NEWS2 combines six physiological parameters—respiratory rate, oxygen saturation, systolic blood pressure, heart rate, temperature, and Glasgow Coma Scale—each categorized and scored from 0 to 3 based on risk tiers. The total score, computed directly from MIMIC-IV data, was incorporated into a univariate logistic regression model for fair comparison. Although NEWS2 achieved a modest AUC of 0.67, its strong calibration and simplicity underline its value as a rapid bedside screening tool rather than a precise mortality predictor.

In summary, the structured models demonstrated strong predictive performance and interpretability. The LASSO and XGBoost models achieved comparable accuracy, with AUCs around 0.75, and provided well-calibrated probability estimates and superior clinical utility relative to NEWS2. While XGBoost slightly improved discrimination, LASSO offered a more parsimonious and easily interpretable framework suitable for clinical deployment. Collectively, these results highlight the advantages of interpretable machine learning models for risk prediction using structured EHR data. The following chapter extends this framework by incorporating unstructured clinical text—such as discharge summaries and radiology reports—to assess whether narrative information further enhances predictive performance.

\chapter{Modeling with Structured and Textual Data}
\label{chap:structured_text_modeling}

This chapter extends the predictive modeling framework by integrating unstructured clinical text—such as discharge summaries and radiology reports—with structured EHR variables to improve in-hospital mortality prediction. Since the theoretical aspects of logistic regression and LASSO were covered earlier, the focus here is on feature selection, model behavior, and the comparative value of textual information.

\section{LASSO Feature Selection with Textual Data}
\label{sec:lasso_textual}

Feature selection in the multimodal feature space was performed using L1-penalized logistic regression with 10-fold cross-validation. The binomial deviance served as the evaluation metric, equivalent to the negative log-likelihood of binary logistic regression. Figure~\ref{app:lasso_cv_structured_textual_deviance} (Appendix) shows the mean deviance across folds as a function of $\log(\lambda)$, highlighting two key regularization parameters: the minimum deviance point ($\lambda_{\text{min}}$) and the 1-SE range.

Compared with the structured-only model, the multimodal configuration exhibited a higher $\lambda_{\text{min}}$ (8.11 vs.\ 1.15), indicating an increased tendency toward overfitting and therefore a need for stronger regularization. The optimal $\lambda_{\text{1SE}}$ remained consistent (24.77), suggesting a similar trade-off between sparsity and model fit. To promote interpretability, we selected the 75th percentile within the 1-SE range, enforcing slightly greater sparsity while maintaining acceptable performance.

\section{Feature Selection in the Multimodal Setting}
\label{sec:feature_selection_multimodal}

After tuning the regularization strength, we fit the final LASSO model and extracted non-zero coefficients. In parallel, an XGBoost classifier was trained on the same dataset, and the top 64 features were selected based on relative importance. Both models identified 64 features, with 30 overlapping predictors spanning vital signs, lab results, and text-derived embeddings—demonstrating strong agreement between linear and non-linear modeling frameworks.

The inclusion of textual features broadened the predictive landscape. TF-IDF and BERT embeddings from discharge summaries appeared among the top predictors, complementing established physiological markers such as Lactate, HR, and GCS\_Total. LASSO emphasized sparse text-based components, while XGBoost highlighted nonlinear interactions and retained more structured variables such as pH, PT, and Glucose. Despite methodological differences, both approaches converged on clinically interpretable predictors.

\section{Univariate Analysis and Collinearity Resolution}
\label{sec:univariate_collinearity}

To refine the multimodal feature set, all variables selected by LASSO and XGBoost were tested via univariate logistic regression. Of the 98 candidate features, 59 were statistically significant (\(p < 0.05\)) predictors of in-hospital mortality, including both structured variables (e.g., Lactate, HR, GCS\_Total) and text-derived embeddings (e.g., disch\_tfidf\_svd\_3, radiology\_bert\_pca\_1).

Multicollinearity was then assessed using Variance Inflation Factors (VIF). Most predictors exhibited acceptable levels (\( \text{VIF} < 5 \)), but several exceeded 10, indicating redundancy—particularly PT and INR, Hemoglobin and Hematocrit, and MBP and DBP. Following standard practice and earlier observations in the structured-only model, we retained PT, Hemoglobin, and MBP, excluding their collinear counterparts to preserve stability and interpretability. A correlation matrix of high-VIF variables is shown in Figure~\ref{app:correlation_matrix_multimodal} (Appendix), confirming these relationships.

\section{Multivariate Logistic Regression in the Multimodal Setting}
\label{sec:logistic_multimodal}

To assess predictive contributions of the selected multimodal features, we fit multivariate logistic regression models using variables identified by both LASSO and XGBoost pipelines. We followed the same stepwise procedure as in the modeling with structured data, but omitted the intermediate results for brevity.

Among the LASSO-selected features, 46 variables were statistically significant in univariate analysis. The resulting multivariate model included several high-impact textual embeddings (e.g., disch\_tfidf\_svd\_1, disch\_tfidf\_svd\_13, discharge\_bert\_pca\_47) alongside structured predictors like Lactate, HR, BUN, and PT. This model achieved a pseudo-\(R^2\) of 0.5226.

Similarly, the XGBoost-selected features yielded a more compact model with 37 variables significant in univariate analysis (after removing collinear ones), producing a logistic model with a pseudo-\(R^2\) of 0.4653. While the explanatory power was slightly lower, the model retained many core clinical variables and highlighted a different subset of textual features.

Finally, after resolving collinearity, we trained a joint multivariate model incorporating all significant variables from both pipelines. This final model included 56 predictors and achieved the best fit, with a pseudo-\(R^2\) of 0.5258. It captured a diverse blend of structured data and embeddings from discharge and radiology notes.

\subsection*{Comparison to Structured-Only Models}

Table~\ref{tab:logistic_comparison_structured_textual} summarizes the logistic regression model performance in structured-only versus multimodal settings.

\begin{table}[htbp]
    \centering
    \begin{tabular}{lccc}
        \toprule
        \textbf{Model} & \textbf{Feature Source} & \textbf{\# Features} & \textbf{Pseudo-\(R^2\)} \\
        \midrule
        LASSO & Structured Only & 16 & 0.1983 \\
        XGBoost & Structured Only & 15 & 0.1957 \\
        \textbf{Combined Model} & \textbf{Structured Only} & \textbf{17} & \textbf{0.1981} \\
        \addlinespace
        LASSO & Structured + Text & 46 & 0.5226 \\
        XGBoost & Structured + Text & 37 & 0.4653 \\
        \textbf{Combined Model} & \textbf{Structured + Text} & \textbf{56} & \textbf{0.5258} \\
        \bottomrule
    \end{tabular}
    \caption{Comparison of logistic regression models in structured-only vs. multimodal settings.}
    \label{tab:logistic_comparison_structured_textual}
\end{table}

These results clearly demonstrate the substantial added value of integrating textual data. All multimodal models substantially outperformed their structured-only counterparts, with pseudo-\(R^2\) improvements ranging from approximately 26 to 33  percentage points. The final model benefits especially from combining interpretable structured features with rich, distributed textual embeddings.




\section{Model Evaluation with Textual Features}
\label{sec:model_evaluation_textual}

To assess the added predictive utility of unstructured clinical text, we compared model performance in terms of ROC curves and decision curve analysis (DCA) across structured-only and multimodal (structured + text) settings.

\subsection*{Discrimination Performance: ROC Curves}

\begin{figure}[htbp]
    \centering
    \includegraphics[width=0.75\textwidth]{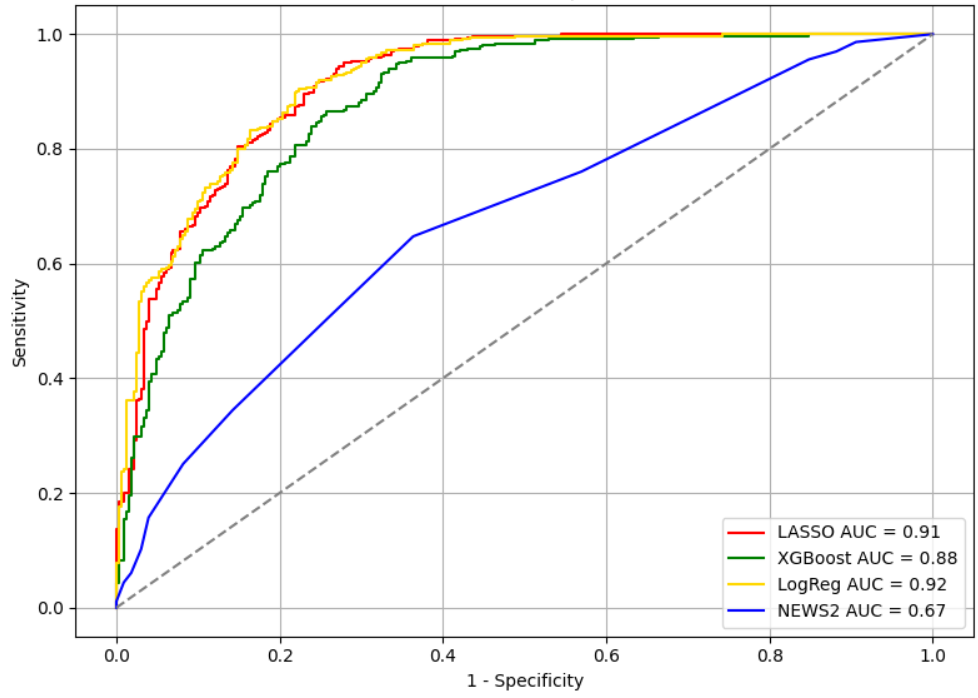}
    \caption{ROC curve comparison after adding textual features.}
    \label{fig:roc_comparison}
\end{figure}

Figure~\ref{fig:roc_comparison} illustrates the ROC curves for all models evaluated on the held-out validation set. Incorporating textual embeddings into the models led to substantial gains in discriminatory power on the validation set. The AUC of the LASSO-based model increased from 0.75 (structured-only) to 0.91 with textual features. Similarly, XGBoost improved from 0.77 to 0.88. The combined logistic regression model, which incorporated structured and text-derived features, achieved the highest AUC of 0.92, outperforming both LASSO and XGBoost. In contrast, the NEWS2 baseline maintained an AUC of 0.67 across both settings, highlighting its limited ability to discriminate outcomes in this ICU cohort.

These results demonstrate that enriching structured clinical data with textual information leads to notable improvements in classification performance, particularly in models that benefit from both expressiveness and interpretability.

\subsection*{Clinical Utility: Decision Curve Analysis}

\begin{figure}[htbp]
    \centering
    \includegraphics[width=0.85\textwidth]{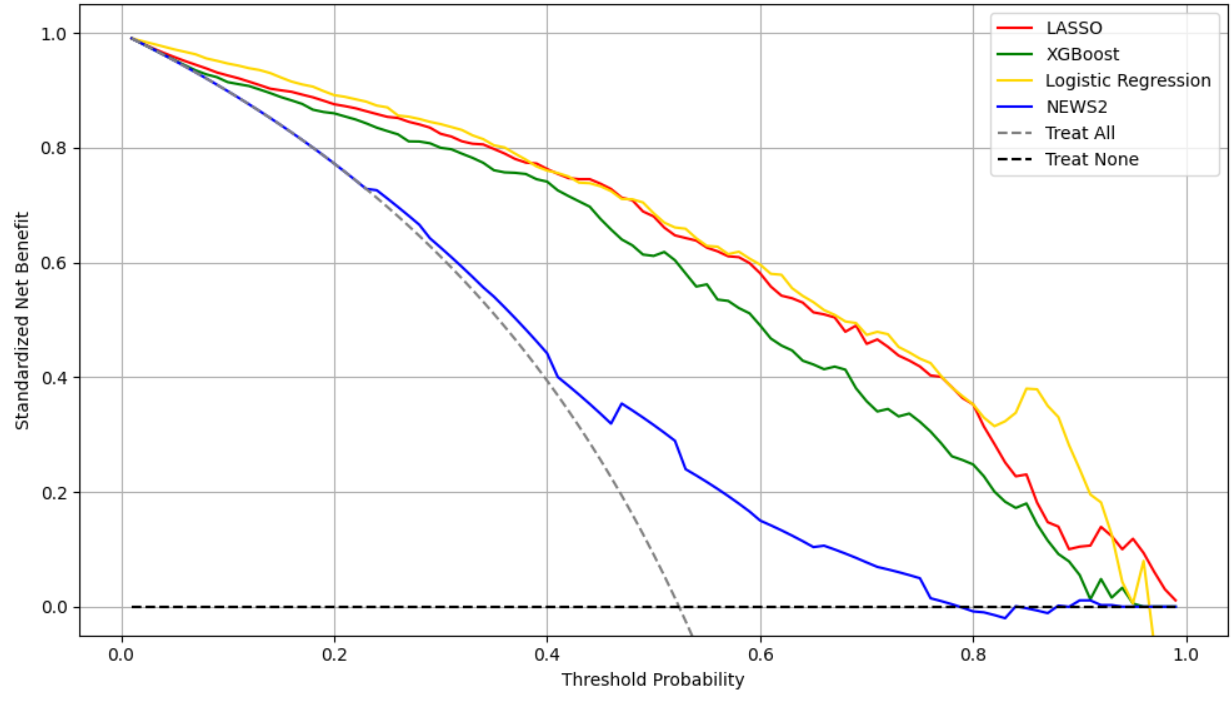}
    \caption{Decision curve analysis comparing structured-only vs. multimodal models.}
    \label{fig:dca_comparison}
\end{figure}

In addition to discrimination, we evaluated clinical utility using decision curve analysis (Figure~\ref{fig:dca_comparison}). Across a broad range of clinically relevant threshold probabilities (approximately 0.2 to 0.8), all multimodal models consistently exhibited higher standardized net benefit than both the structured-only variants and the default strategies of treating all or treating none.

Among the models, logistic regression achieved the highest net benefit across most thresholds, narrowly surpassing LASSO and clearly outperforming XGBoost. These findings suggest that interpretable linear models, when augmented with informative textual embeddings, can rival—or even surpass—more complex non-linear approaches in actionable clinical value. The inclusion of discharge summary and radiology report embeddings appears to provide signal that enhances both model discrimination and clinical decision benefit.

\subsection*{Summary of Evaluation Improvements}

Table~\ref{tab:model_eval_metrics} summarizes the key evaluation metrics on the validation set for the top-performing structured-only and multimodal models.

\begin{table}[htbp]
    \centering
    \begin{tabular}{lcc}
        \toprule
        \textbf{Metric} & \textbf{Structured Only} & \textbf{Structured + Text} \\
        \midrule
        AUC & 0.7534 & \textbf{0.9180} \\
        Accuracy & 0.68 & \textbf{0.84} \\
        F1-score (Class 1) & 0.70 & \textbf{0.85} \\
        Recall (Class 1) & 0.71 & \textbf{0.88} \\
        \bottomrule
    \end{tabular}
    \caption{Validation performance metrics comparing structured-only and multimodal models.}
    \label{tab:model_eval_metrics}
\end{table}

Across all metrics—AUC, accuracy, F1-score, and recall—substantial improvements were observed with the addition of textual features. The multimodal model’s F1-score for the positive class increased from 0.70 to 0.85, and recall rose from 0.71 to 0.88, demonstrating stronger detection of high-risk patients.

These gains indicate that textual embeddings contribute meaningful clinical context that structured variables alone cannot capture. In particular, higher recall is crucial in ICU mortality prediction, where identifying high-risk patients early outweighs the cost of false positives. Overall, the integration of unstructured clinical narratives markedly enhanced both statistical performance and potential clinical utility.


\chapter{Conclusion}
\label{chap:conclusion}

This research examined how combining structured electronic health record (EHR) variables with unstructured clinical text can improve early prediction of in-hospital mortality among intensive care patients. Using data from the MIMIC-IV database, we integrated physiological measurements, laboratory results, and narrative reports such as discharge summaries and radiology notes into a unified modeling framework. Textual information was transformed using TF-IDF and BioBERT embeddings, reduced through SVD and PCA, and merged with structured features for multimodal learning.

Models based solely on structured data achieved moderate performance (AUC $\approx$ 0.75). Incorporating textual features significantly enhanced discrimination and recall, with the best multimodal model reaching an AUC of 0.92. These results demonstrate that free-text clinical documentation contains complementary prognostic information not captured in structured data and that interpretable models like LASSO-regularized logistic regression can effectively exploit this information without sacrificing transparency.

The findings highlight the potential of multimodal EHR modeling to improve mortality risk prediction and support clinical decision-making. Although the analysis was based on data from a single institution and post hoc notes, the framework can be readily extended to other hospitals and real-time contexts. Future work should focus on temporal modeling of early clinical narratives, external validation across healthcare systems, and improving interpretability of text-derived features.

Overall, this study provides empirical evidence that integrating structured and unstructured data enhances predictive performance and clinical insight, supporting the development of more accurate and explainable decision-support tools in critical care.

\bibliography{references}

\cleardoublepage
\appendix
\appendixpage

\chapter{Supplementary Tables}
\label{app:missing_data}

\begin{table}[h]
\centering
\caption{Missing Values Summary (Vital Signs)}
\label{tab:missing_vitals}
\begin{tabular}{lcc}
\toprule
\textbf{Vital Sign} & \textbf{Missing Count} & \textbf{Missing \%} \\
\midrule
HR & 6 & 0.26\% \\
SBP & 35 & 1.52\% \\
DBP & 38 & 1.65\% \\
MBP & 38 & 1.65\% \\
RR & 14 & 0.61\% \\
BT & 307 & 13.31\% \\
SpO$_2$ & 36 & 1.56\% \\
\bottomrule
\end{tabular}
\end{table}

\begin{table}[h]
\centering
\caption{Missing Values Summary (Lab Tests)}
\label{tab:missing_labs}
\begin{tabular}{lcc}
\toprule
\textbf{Lab Test} & \textbf{Missing Count} & \textbf{Missing \%} \\
\midrule
Hematocrit & 106 & 4.59\% \\
Hemoglobin & 112 & 4.85\% \\
Platelet & 117 & 5.07\% \\
WBC & 125 & 5.42\% \\
PT & 252 & 10.92\% \\
INR & 249 & 10.79\% \\
Creatinine & 102 & 4.42\% \\
BUN & 107 & 4.64\% \\
Glucose & 110 & 4.77\% \\
Potassium & 104 & 4.51\% \\
Sodium & 97 & 4.20\% \\
Calcium & 157 & 6.81\% \\
Chloride & 99 & 4.29\% \\
Anion Gap & 102 & 4.42\% \\
Bicarbonate & 101 & 4.38\% \\
Lactate & 442 & 19.16\% \\
pH & 406 & 17.60\% \\
\bottomrule
\end{tabular}
\end{table}

\label{app:vif_table}
\begin{table}[ht]
\centering
\caption{Final VIF Scores After Dropping \texttt{INR}}
\label{table:vif_scores}
\begin{tabular}{lr}
\toprule
\textbf{Variable} & \textbf{VIF} \\
\midrule
Lactate & 3.32 \\
SBP & 1.18 \\
BUN & 1.43 \\
Bicarbonate & 2.07 \\
HR & 1.38 \\
RR & 1.40 \\
COPD & 1.05 \\
PT & 1.25 \\
Hemoglobin & 1.26 \\
WBC & 1.14 \\
SpO\textsubscript{2} & 1.29 \\
GCS\_Total & 1.35 \\
pH & 2.29 \\
BT & 1.34 \\
anchor\_age & 1.16 \\
AnionGap & 2.76 \\
congestive\_heart\_failure & 1.15 \\
\bottomrule
\end{tabular}
\end{table}

\chapter{Supplementary Figures}
\label{app:vital_figures}

\begin{figure}[h]
    \centering
    \includegraphics[width=\textwidth]{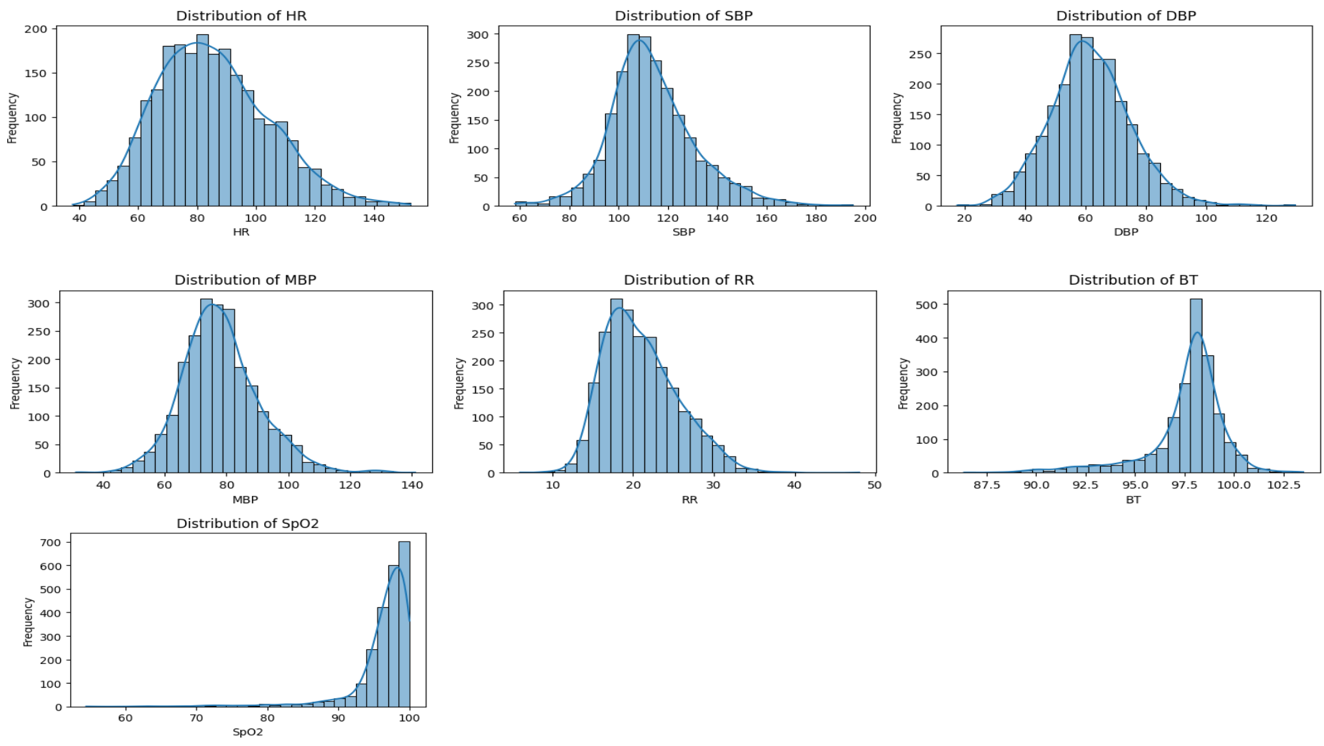}
    \caption{Distribution of Vital Signs across the first 24 hours of ICU admission.}
    \label{fig:vital_distributions}
\end{figure}

\label{app:lab_figures}
\begin{figure}[h]
    \centering
    \includegraphics[width=1.0\textwidth, height=7.5cm]{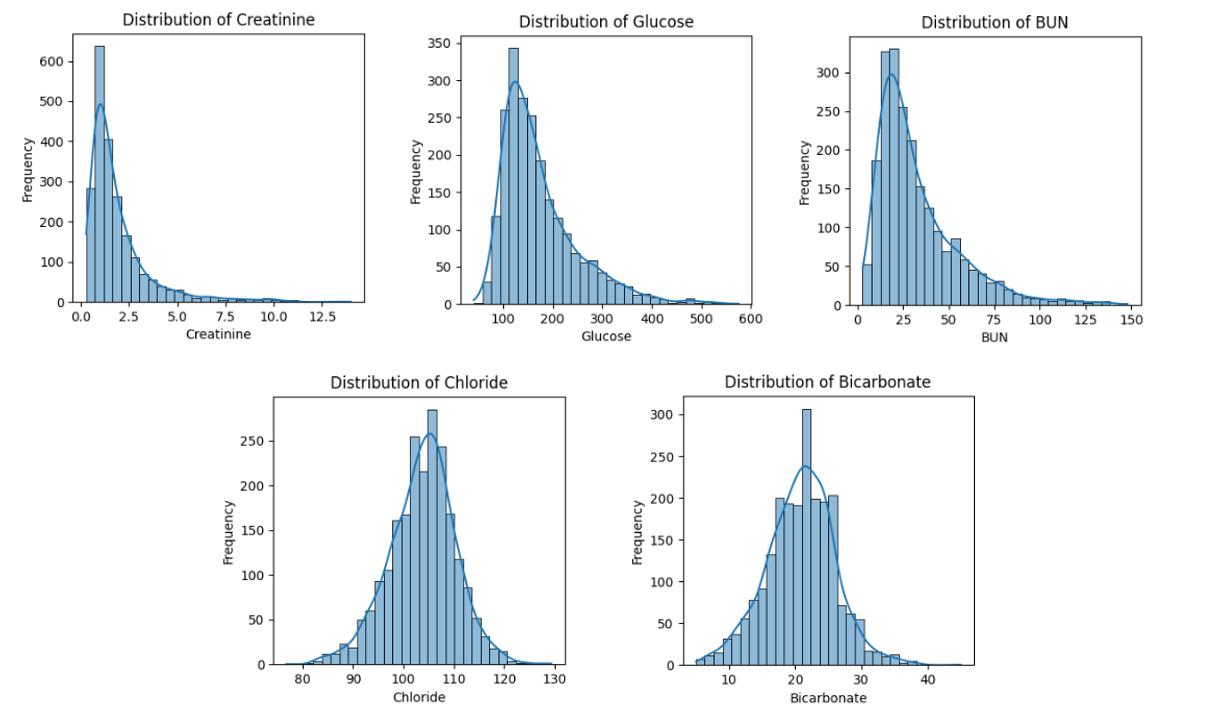}
    \caption{Distributions of Lab Test Values}
    \label{fig:lab_distributions}
\end{figure}

\label{app:gcs_figures}
\begin{figure}[h]
    \centering
    \includegraphics[width=1.0\textwidth]{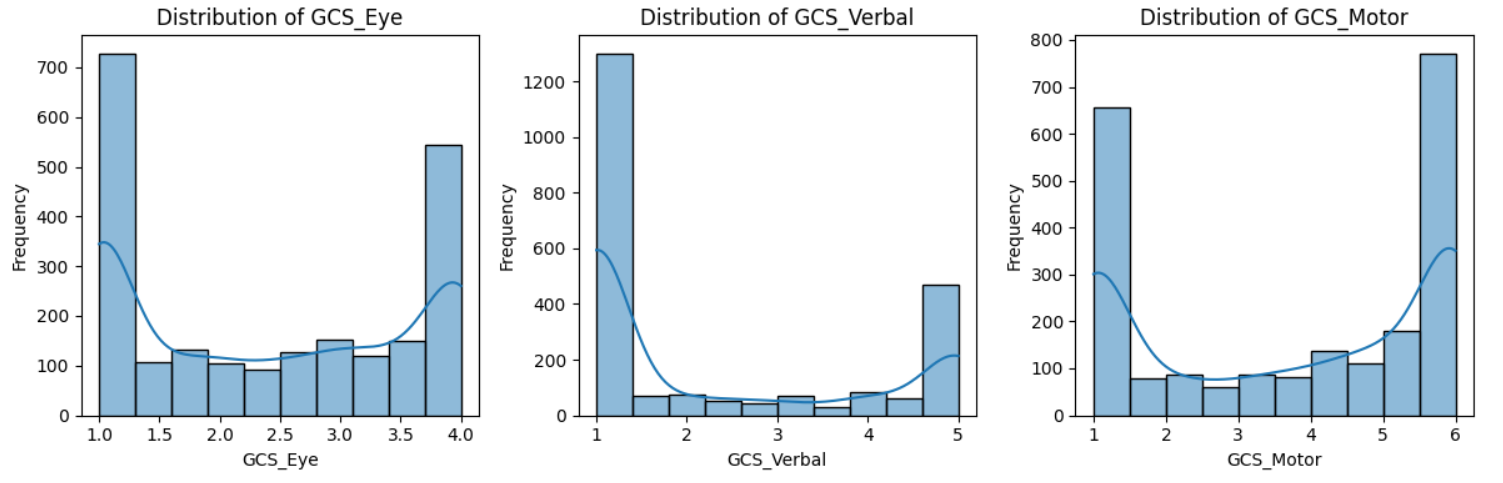}
    \caption{Distribution of GCS Eye, Verbal, and Motor components across ICU patients.}
    \label{fig:gcs_distributions}
\end{figure}

\label{app:text_figures_tfidf}
\begin{figure}[h]
    \centering
    \begin{subfigure}[b]{0.48\textwidth}
        \centering
        \includegraphics[height=4.05cm]{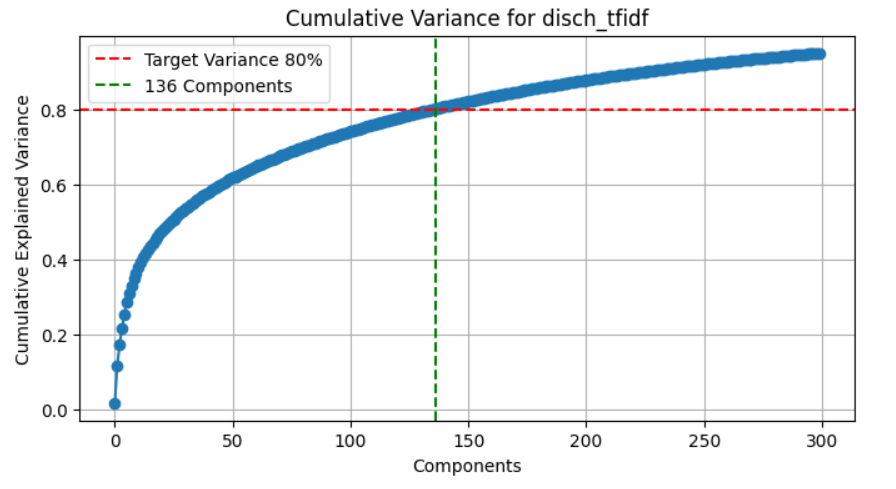}
        \caption{Discharge TF-IDF}
        \label{fig:svd_disch}
    \end{subfigure}
    \hfill
    \begin{subfigure}[b]{0.48\textwidth}
        \centering
        \includegraphics[height=4.05cm]{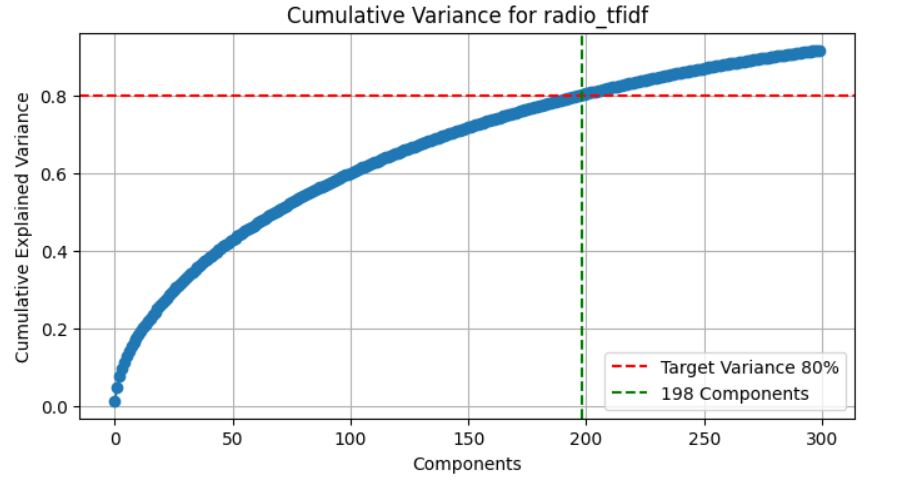}
        \caption{Radiology TF-IDF}
        \label{fig:svd_radio}
    \end{subfigure}
    \caption{Cumulative variance explained by SVD components}
    \label{fig:svd_combined}
\end{figure}

\label{app:text_figures_bert}
\begin{figure}[h]
    \centering
    \begin{subfigure}[b]{0.48\textwidth}
        \centering
        \includegraphics[height=4.05cm]{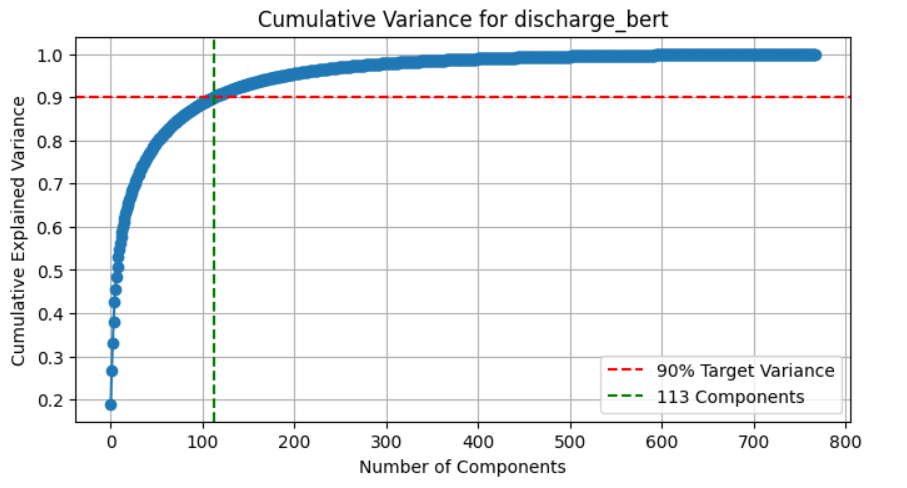}
        \caption{Discharge embeddings}
        \label{fig:bert_disch}
    \end{subfigure}
    \hfill
    \begin{subfigure}[b]{0.48\textwidth}
        \centering
        \includegraphics[height=4.05cm]{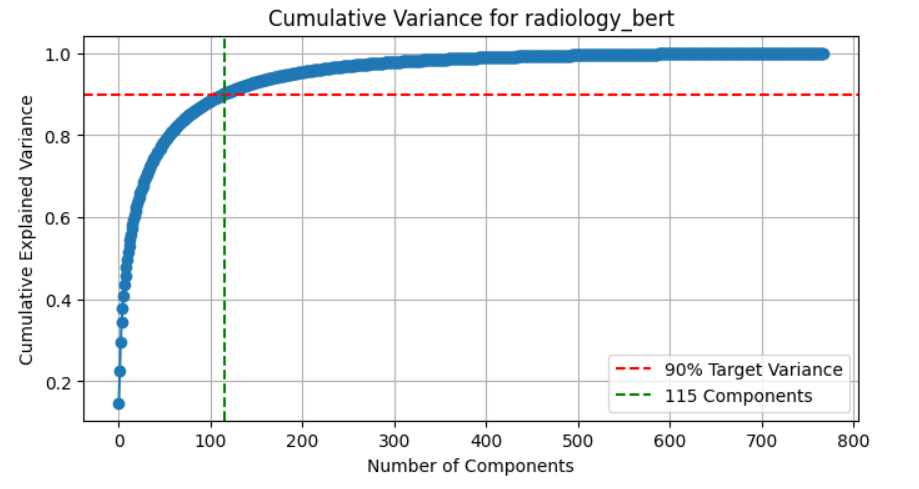}
        \caption{Radiology embeddings}
        \label{fig:bert_radio}
    \end{subfigure}
    \caption{Cumulative explained variance for PCA on BioBERT embeddings}
    \label{fig:bert_combined}
\end{figure}

\label{app:roc_structured}
\begin{figure}[htbp]
    \centering
    \includegraphics[width=0.75\textwidth]{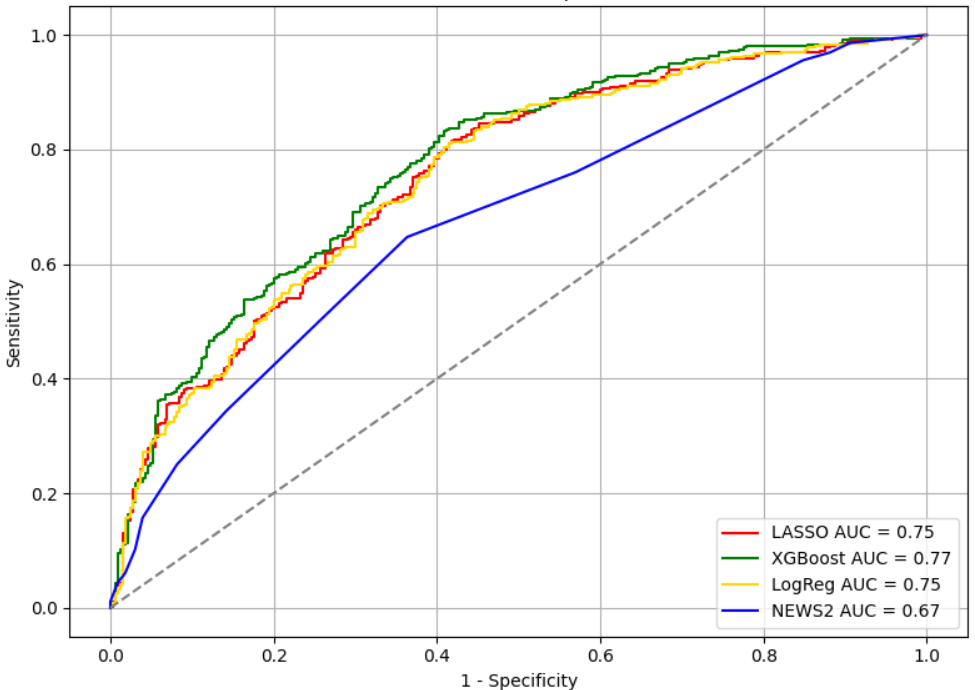}
    \caption{ROC curves comparing models: LASSO, XGBoost, Logistic Regression, and NEWS2. XGBoost achieved the highest AUC on the validation set.}
    \label{fig:roc_structured_models}
\end{figure}

\label{app:calibration_structured_models}
\begin{figure}[htbp]
    \centering
    \begin{subfigure}[b]{0.32\textwidth}
        \includegraphics[width=\textwidth, height=5.0cm]{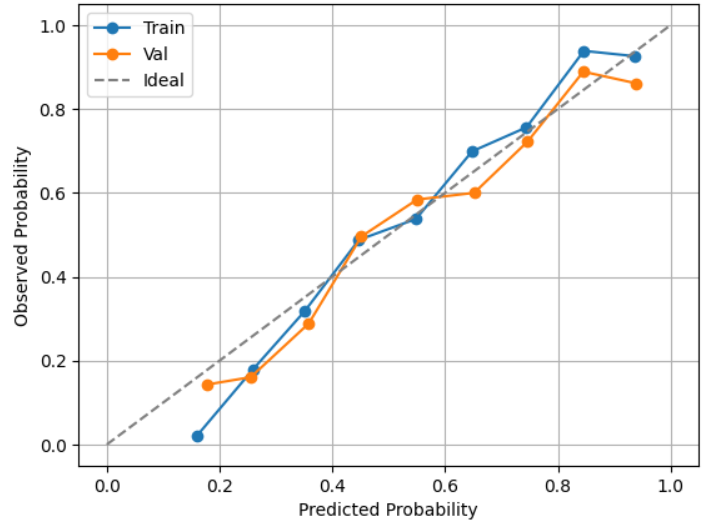}
        \caption{LASSO}
        \label{fig:calibration_lasso}
    \end{subfigure}
    \hfill
    \begin{subfigure}[b]{0.32\textwidth}
        \includegraphics[width=\textwidth, height=5.0cm]{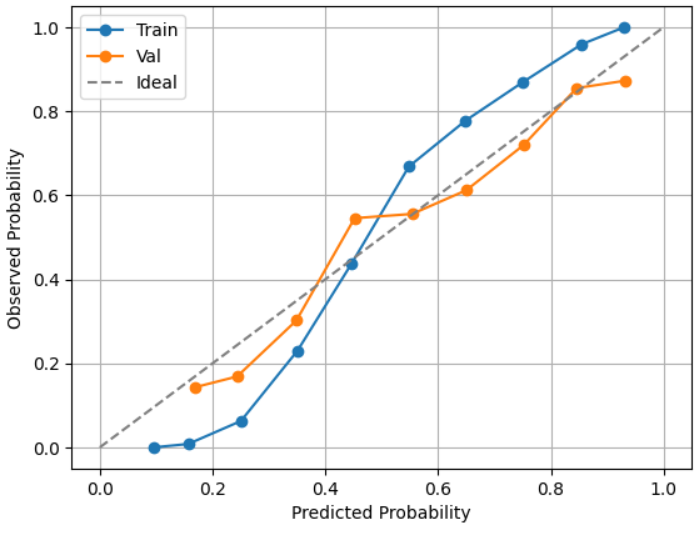}
        \caption{XGBoost}
        \label{fig:calibration_xgboost}
    \end{subfigure}
    \hfill
    \begin{subfigure}[b]{0.32\textwidth}
        \includegraphics[width=\textwidth, height=5.0cm]{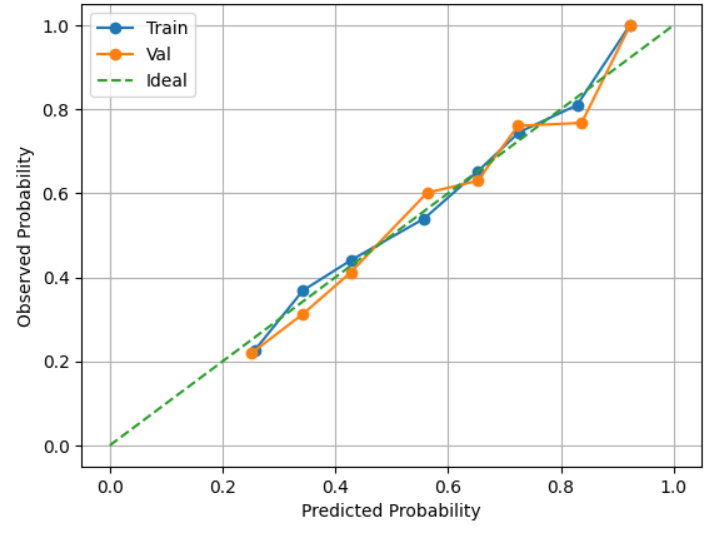}
        \caption{NEWS2}
        \label{fig:calibration_news2}
    \end{subfigure}
    \caption{Calibration plots for LASSO, XGBoost, and NEWS2 models on the validation set. Closer alignment with the diagonal line indicates better probability calibration.}
    \label{fig:calibration_structured_models}
\end{figure}

\label{app:dca_structured_models}
\begin{figure}[htbp]
    \centering
    \includegraphics[width=0.75\textwidth]{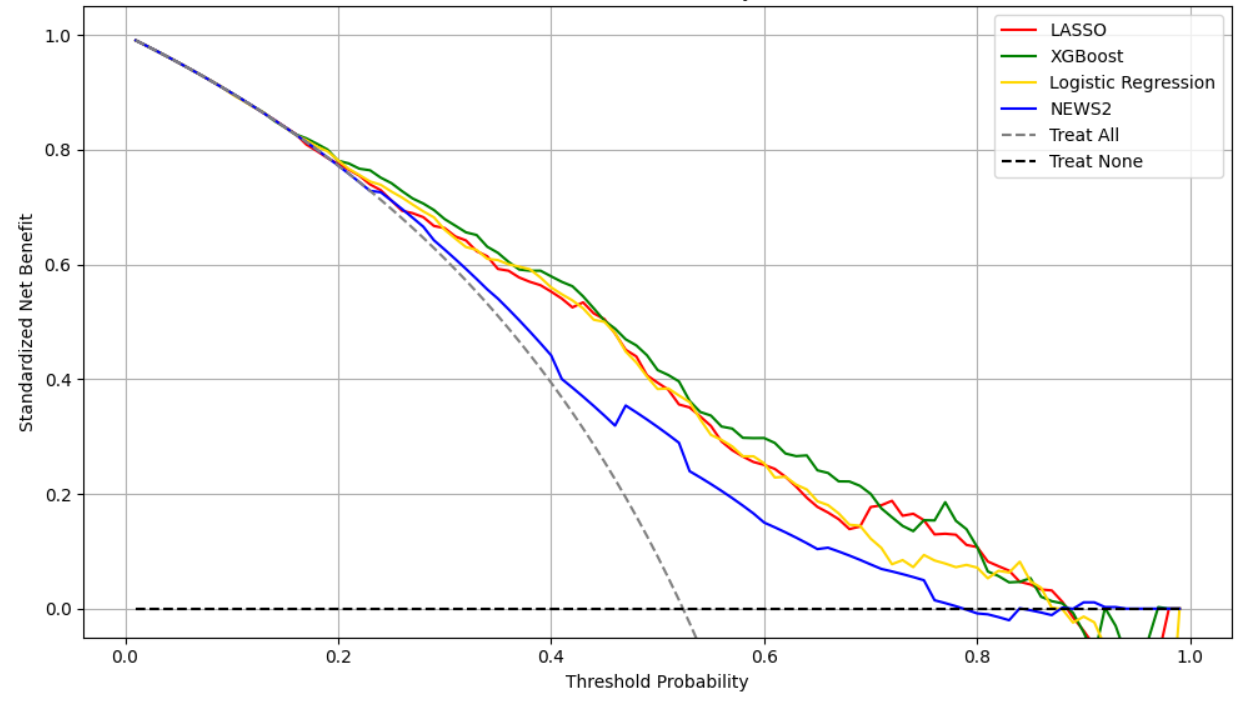}
    \caption{Decision Curve Analysis comparing net clinical benefit of structured models. LASSO and XGBoost achieved the highest net benefit across a broad range of threshold probabilities.}
    \label{fig:dca_structured_models}
\end{figure}

\label{app:lasso_cv_structured_textual_deviance}
\begin{figure}[htbp]
    \centering

    \begin{subfigure}[b]{0.48\textwidth}
        \includegraphics[width=\textwidth]{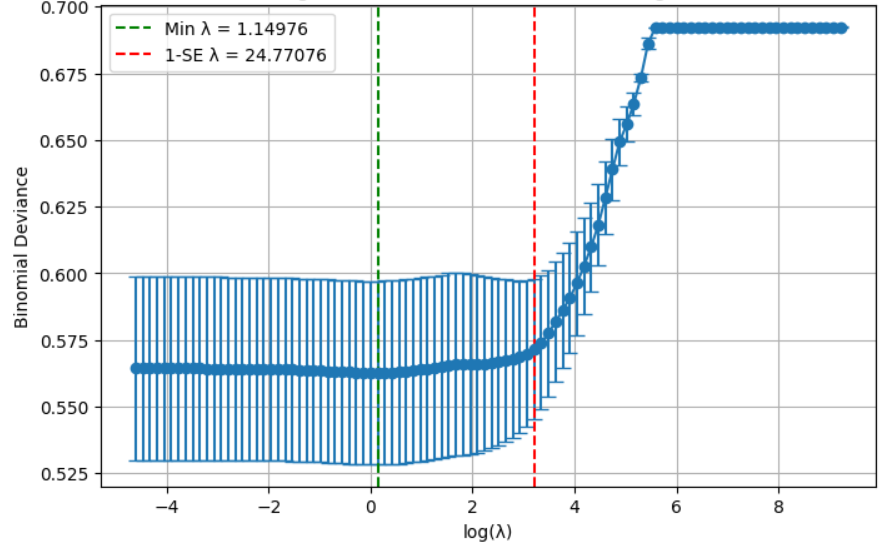}
        \caption{Structured features only}
        \label{fig:lasso_cv_deviance_structured}
    \end{subfigure}
    \hfill
    \begin{subfigure}[b]{0.48\textwidth}
        \includegraphics[width=\textwidth]{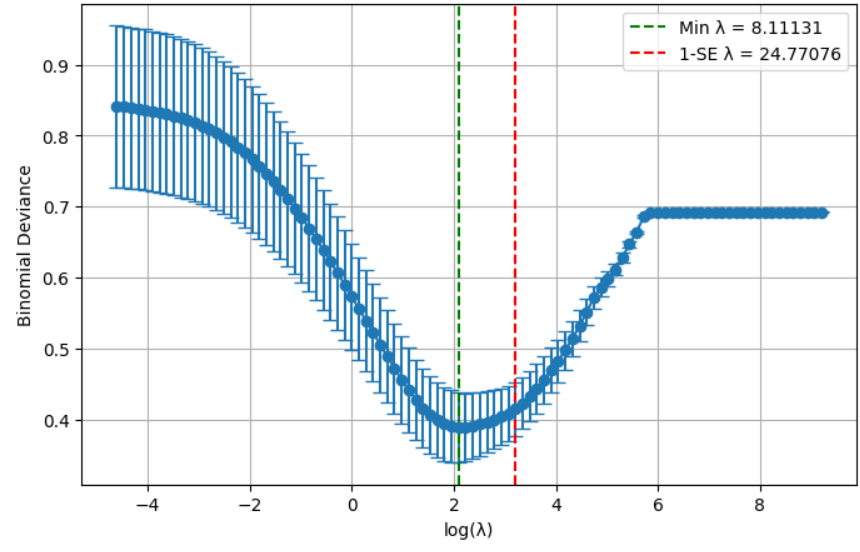}
        \caption{Structured + textual features}
        \label{fig:lasso_cv_deviance_textual}
    \end{subfigure}

    \caption{LASSO cross-validation binomial deviance curves for feature selection. Vertical dashed lines indicate \(\lambda_{\text{min}}\) and \(\lambda_{\text{1SE}}\). Incorporating textual features shifts the optimal \(\lambda\) values, reflecting improved fit.}
    \label{fig:lasso_cv_deviance_comparison}
\end{figure}

\label{app:correlation_matrix_multimodal}
\begin{figure}[h]
    \centering
    \includegraphics[width=0.8\textwidth]{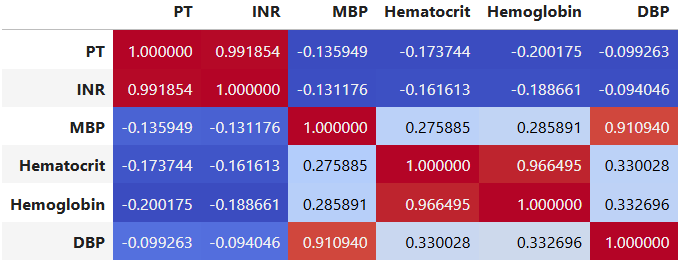}
    \caption{Correlation matrix among high-VIF variables in the multimodal setting.}
    \label{fig:correlation_matrix_multimodal}
\end{figure}

\end{document}